\documentclass[runningheads]{llncs}

\usepackage[T1]{fontenc}
\usepackage{graphicx,verbatim}
\usepackage{caption}
\usepackage{hyperref}
\usepackage{multirow}
\usepackage[table]{xcolor}
\usepackage{pifont}
\usepackage{amsmath}
\usepackage{booktabs}
\usepackage{orcidlink}

\newcommand*\colourcheck[1]{%
  \expandafter\newcommand\csname #1check\endcsname{\textcolor{#1}{\ding{52}}}%
}
\colourcheck{green}
\newcommand*\colourcross[1]{%
  \expandafter\newcommand\csname #1cross\endcsname{\textcolor{#1}{\ding{56}}}%
}
\colourcross{red}

\newcommand\tablefontsize{\fontsize{8pt}{8pt}\selectfont}
\usepackage{color}

\begin{document}

\title{Zero-shot Monocular Metric Depth for Endoscopic Images}

\author{
Nicolas Toussaint\inst{1}\orcidlink{0000-0003-4696-465X} \and
Emanuele Colleoni\inst{1}\orcidlink{0000-0003-4614-5742} \and
Ricardo Sanchez-Matilla\inst{1}\orcidlink{0000-0003-2330-0973} \and
Joshua Sutcliffe\inst{1} \and
Vanessa Thompson\inst{1} \and
Muhammad Asad\inst{1}\orcidlink{0000-0002-3672-2414} \and
Imanol Luengo\inst{1}\orcidlink{0000-0001-6573-322X} \and
Danail Stoyanov\inst{1,2}\orcidlink{0000-0002-0980-3227}
}

\authorrunning{N. Toussaint et al.}
\institute{
Medtronic Digital Technologies, London, UK \and
UCL Hawkes Institute, University College London, London, UK \\
\email{nicolas.toussaint@medtronic.com}
}

\maketitle

\begin{abstract}
Monocular relative and metric depth estimation has seen a tremendous boost in the last few years due to the sharp advancements in foundation models and in particular transformer based networks. As we start to see applications to the domain of endoscopic images, there is still a lack of robust benchmarks and high-quality datasets in that area. This paper addresses these limitations by presenting a comprehensive benchmark of state-of-the-art (metric and relative) depth estimation models evaluated on real, unseen endoscopic images, providing critical insights into their generalisation and performance in clinical scenarios. Additionally, we introduce and publish a novel synthetic dataset (EndoSynth) of endoscopic surgical instruments paired with ground truth metric depth and segmentation masks, designed to bridge the gap between synthetic and real-world data. We demonstrate that fine-tuning depth foundation models using our synthetic dataset boosts accuracy on most unseen real data by a significant margin. By providing both a benchmark and a synthetic dataset, this work advances the field of depth estimation for endoscopic images and serves as an important resource for future research. Project page, EndoSynth dataset and trained weights are available at \href{https://github.com/TouchSurgery/EndoSynth}{https://github.com/TouchSurgery/EndoSynth}
\keywords{Minimally Invasive Surgery \and Depth Estimation \and Public Dataset}

\end{abstract}

\section{Introduction}

Monocular Depth Estimation (MDE) has seen significant improvements in recent years due to the rise of foundation models, large-scale, pretrained deep neural networks designed to serve as versatile bases for a wide range of downstream tasks with minimal fine-tuning. In particular, there has been a focus on models based on Vision Transformer (ViT) encoders, such as DepthAnything~\cite{yang2024depth}~(DAv1) and DepthAnythingV2~\cite{yang2024depthv2}~(DAv2), as ViTs have emerged as very strong backbone encoders for downstream tasks~\cite{kiyasseh2023vision}. Transferring these foundation models to the domain of surgical vision is surfacing~\cite{cui2024endodac,budd2024transferring,han2024depth}. This transfer faces two main challenges (i) surgical images exhibit different depth distributions compared to natural image datasets, (ii) endoscope camera intrinsics differ from standard devices (e.g. phone cameras).
\begin{figure}[t]
\centering
\includegraphics[width=.98\textwidth]{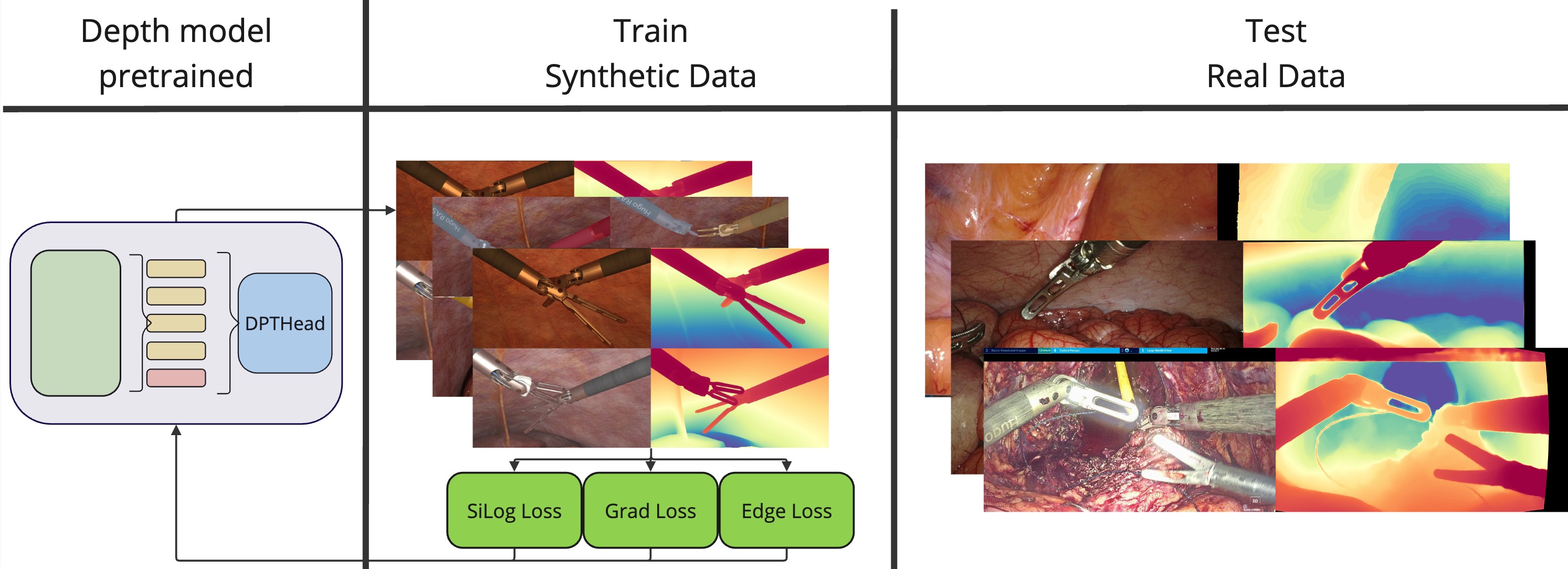}
\caption{Overview: Models are fine-tuned using EndoSynth, our contributed synthetic depth dataset, and evaluated on real unseen surgical frames from several datasets.}
\label{fig:overview}
\end{figure}
Best performing networks in MDE often use synthetic labels to boost training data. We note for instance DAv2 using ~500k synthetic images as their first stage of training.
Simulation engines, based on advancements in computer graphics can offer precise annotations crucial for this purpose. Prior studies have explored training deep learning models using synthetic images. However these typically rely on complex style-transfer algorithms~\cite{rivoir2021long,walkner2024synthetic}.
To address these challenges, this paper proposes the following contributions:

\begin{itemize} 
\item \textbf{Benchmark of state-of-the-art depth estimation models:} We provide a comprehensive evaluation of leading depth estimation models on several publicly available endoscopic datasets as well as on an in-house clinical dataset for the task of metric depth estimation. We deliver critical insights into their generalisability and performance. This study aims to identify the most effective models currently available, particularly in the clinical setting, serving as a valuable reference for future research in the field.

\item \textbf{Novel synthetic dataset of endoscopic surgical instruments with ground truth metric depth:} We introduce and release EndoSynth, a new synthetic dataset featuring endoscopic Hugo\textsuperscript{TM} RAS surgical instruments paired with precise ground truth metric depth and segmentation masks. This dataset addresses the scarcity of high-quality training data and aims to bridge the gap between synthetic and real-world endoscopic imaging. We perform meaningful experiments showcasing the effectiveness of fine-tuning state-of-the-art models on our dataset for the task of metric depth estimation and instrument semantic segmentation.
\end{itemize}

\section{Material and Methods}

\subsection{Existing endoscopic depth datasets}

We summarise the endoscopic datasets used in this work in Table~\ref{tab:datasets}. SCARED~\cite{allan2021stereo} comprises 45 frames with sparse ground truth depth estimated by structured light. Following~\cite{budd2024transferring}, we exclude additional frames as they are propagations of the first frames. EndoNERF~\cite{wang2022neural} dataset consists of 2 sequences of 156 and 63 frames each. We select four clips from P1, P2\_0, P2\_6 and P2\_7 from StereoMIS~\cite{hayoz2023learning} due to their higher diversity and motion. We extract frames at 1~fps. Rectified-Hamlyn~\cite{ye2017self} comprises 4 in-vivo clips of rectified images. We follow~\cite{recasens2021endo} and use sequences 1,4,19,20, from which we randomly select 200 frames in each.

\subsection{EndoSynth: Depth dataset for endoscopic surgical instruments}

\begin{figure}[t]
\includegraphics[height=.38\textwidth]{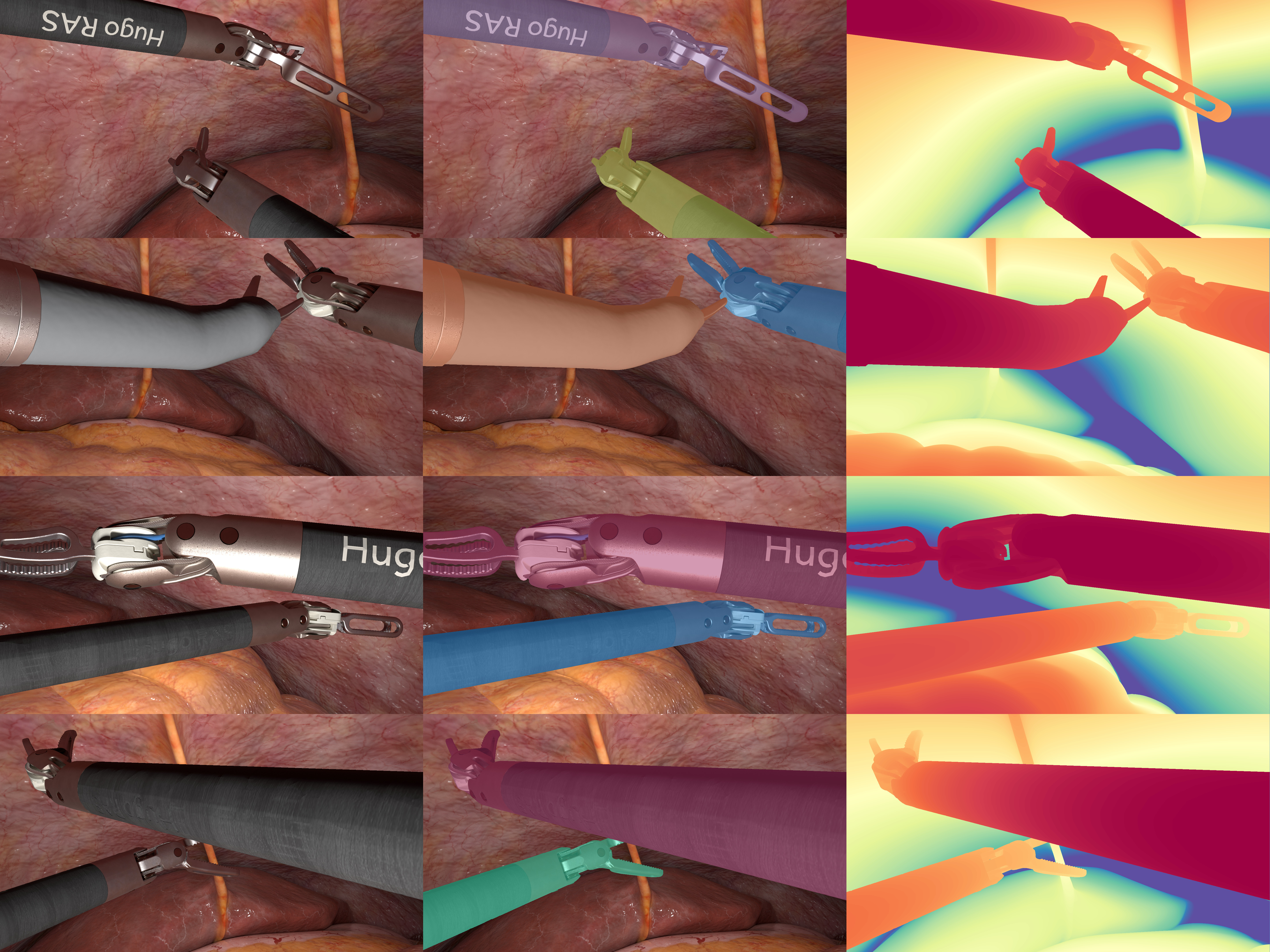}
\includegraphics[height=.39\textwidth]{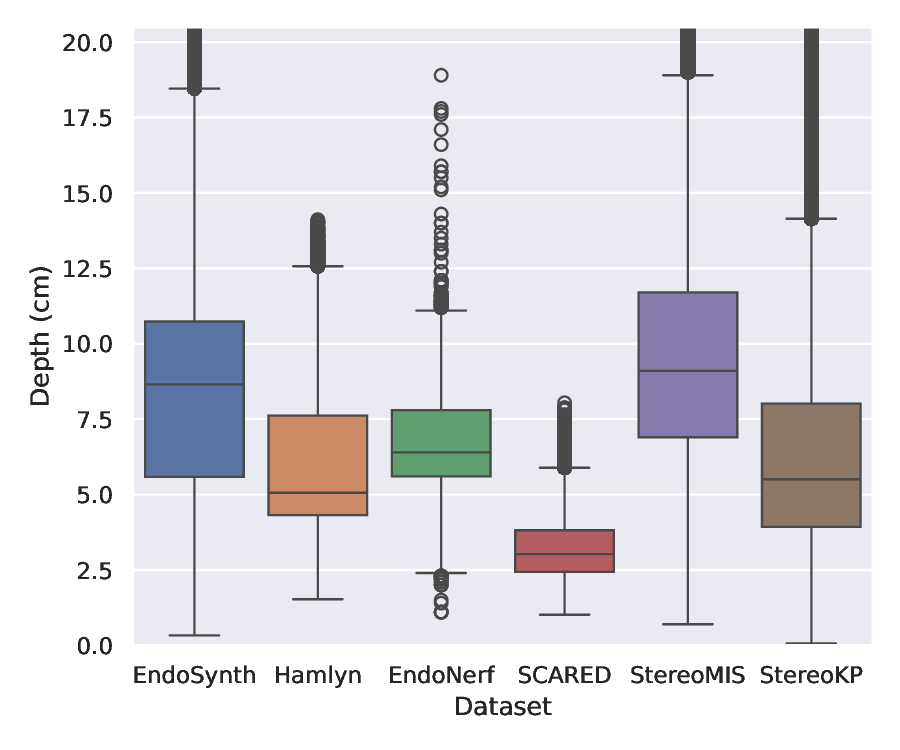}
\caption{EndoSynth samples. Left to right: RGB, instrument segmentation, metric depth (blue: farther / red: closer), ranges of depths observed in EndoSynth and other public datasets.}
\label{fig:endosynth_samples}
\end{figure}

We release a novel synthetic dataset dedicated to endoscopic surgical images. The dataset is composed of 2,000 synthetically generated frames, mimicking the human abdominal cavity, and Hugo\textsuperscript{TM} RAS~\cite{ngu2024narrative} robotic surgical instruments in the foreground. 

\subsubsection*{Synthetic data generation:} 

\begin{itemize}
    \item Background: a static 3D scene mimicking the abdominal cavity was produced using the 3D rendering software Maya\textsuperscript{\textcopyright}. 
    \item Foreground: Computer-Assisted Design (CAD) models of four Hugo\textsuperscript{TM} compatible surgical instruments were used. The models were randomly placed at plausible locations between the camera and the background scene.
    \item Camera: All frames share the same intrinsic camera parameters, resembling the camera parameters observed in Hugo\textsuperscript{TM} compatible endoscopic cameras.
    \item Rendering: The ray-tracing rendering was performed using Maya's V-Ray\textsuperscript{\textcopyright} plugin. Each frame is accompanied by a metric depth map and pixel-wise segmentation labels for the surgical instruments.
\end{itemize}

Samples of the contributed EndoSynth dataset are presented in Fig.~\ref{fig:endosynth_samples}. We show the ranges of observed depth in different publicly available datasets compared to EndoSynth in Fig.~\ref{fig:endosynth_samples} (right). We notice from this plot that SCARED ranges is significantly lower than its counterparts.

\subsubsection*{StereoKP:} 

We further extend our benchmarks to an in-house dataset. It comprises 1,500 stereo-frames from three videos of minimally invasive surgery on patients, two partial nephrectomy procedures and one radial nephrectomy. Keypoints of surgical instruments were manually annotated in both left and right frames, and precise 3D localization of the keypoints were obtained via triangulation. StereoKP provides a significant additional insight to the ability of depth models to accurately estimate metric depth on instruments.

\begin{table}[t]
\centering
\caption{Endoscopic datasets used in this work. \greencheck indicates that the depth ground truth is non-derived from AI models.}
\label{tab:datasets}
\begin{tabular}{|l|l|r|r|r|c|}
\toprule
\textbf{Dataset} & \textbf{Domain} & \textbf{Depth source} & \textbf{Frames} & \textbf{Clips} & \textbf{Usage} \\
\midrule
EndoSynth* (ours) & abdom. cavity & \greencheck generated & 2,000 & N/A & Train  \\
\midrule
Hamlyn~\cite{ye2017self} & in-vivo mixed & stereo-match & 800 & 4 & Test  \\
EndoNERF~\cite{wang2022neural} & in-vivo porcine & stereo-match & 156 & 2 & Test  \\
SCARED~\cite{allan2021stereo} & ex-vivo porcine & \greencheck structured-light & 45 & 45 & Test \\
StereoMIS~\cite{hayoz2023learning} & in-vivo porcine & stereo-match & 474 & 4 & Test  \\
StereoKP** & in-vivo porcine & \greencheck annotated & 1,500 & 3 & Test  \\
\bottomrule
\multicolumn{4}{l}{\small *dataset released with this work\footnotemark{}, **in-house dataset}
\end{tabular}
\end{table}

\footnotetext{EndoSynth dataset linked at \href{https://github.com/TouchSurgery/EndoSynth}{https://github.com/TouchSurgery/EndoSynth}}

\subsection{Models}

Recent advances in depth estimation predominantly leverage ViT-based encoder backbones. DAv1 and DAv2 employ a DinoV2-pretrained~\cite{oquab2023dinov2} ViT encoder coupled with a Depth Prediction Transformer~\cite{ranftl2021vision} (DPT) head. DepthPro~\cite{bochkovskii2024depth} integrates dual full-size ViT encoders (for patch and image-level features) alongside a third ViT for field-of-view regression. EndoDAC~\cite{cui2024endodac} augments both the encoder and DPT head with additional architectural blocks, while Budd~et~al.~\cite{budd2024transferring} introduce self-supervision consistency losses into the training framework. Midas-v3.1~\cite{birkl2023midas} finds best results using a BEiT backbone paired with a DPTHead.

In this study, we benchmark four models: DAv1, DAv2, EndoDAC (all using base-sized configurations), and Midas (v3.1, BEiT-l-512). All models apart from DAv2 were originally trained to predict inverse depth. To enable direct metric depth estimation, we modify their final activation functions, aligning their outputs to produce physically meaningful depth values:

\begin{equation}
    d = 
    \begin{cases}
      d_{max} \cdot \sigma \left( - \log \left(\text{ReLU}(x) + \epsilon \right) \right) &\text{ for DAv1} \\
      d_{max} \cdot \sigma \left( \text{ReLU}(x) \right) &\text{ for DAv2} \\
      1 / \left(\Delta_{min} + \Delta_{range} \cdot x \right) &\text{ for EndoDAC and Midas}
    \end{cases}
    \label{eq:act}
\end{equation}

\noindent where $d_{max}$ is the maximum depth which is set to 0.3 meters, $\sigma$ is the sigmoid function, $\Delta_{min}, \Delta_{range}$ is the minimum and range of disparities, and $\epsilon$ is a small number ensuring numerical stability. For DAv2, we start from the metric depth version of the model, so the activation reduces to a simple sigmoid function following ReLU.

\begin{table}[tb]
\tablefontsize

\centering
\caption{Zero-shot \emph{relative} depth performances on each test dataset. "-F": models fine-tuned using EndoSynth only (\textbf{best}, \underline{second best}, \colorbox{green!20}{improved}).}
\label{tab:eval_reldepth}
\begin{tabular}{|c|l|ll|ll|ll|ll|}
\toprule
 & Model & DAv1 & DAv1-F & DAv2 & DAv2-F & EndoDAC & EndoDAC-F & Midas & Midas-F \\
Metric & Dataset &  &  &  &  &  &  &  &  \\
\midrule
\multirow[c]{5}{*}{AbsRel $\downarrow$} 
 & Hamlyn & 0.324 & \cellcolor{green!20}0.304 & 0.312 & \cellcolor{green!20}\underline{0.301} & 0.310 & \cellcolor{green!20}\textbf{0.298} & 0.510 & \cellcolor{green!20}0.320 \\
 & EndoNerf & 0.261 & \cellcolor{green!20}0.147 & 0.173 & \cellcolor{green!20}\textbf{0.143} & 0.191 & \cellcolor{green!20}\underline{0.146} & 0.998 & \cellcolor{green!20}0.146 \\
 & SCARED & 0.265 & \cellcolor{green!20}\underline{0.241} & 0.250 & \cellcolor{green!20}0.246 & \textbf{0.230} & 0.249 & 0.548 & \cellcolor{green!20}0.262 \\
 & StereoMIS & 0.238 & \cellcolor{green!20}0.209 & \underline{0.205} & \cellcolor{green!20}\textbf{0.203} & 0.223 & 0.239 & 0.660 & \cellcolor{green!20}0.249 \\
 & StereoKP & 0.244 & \cellcolor{green!20}\underline{0.153} & 0.246 & \cellcolor{green!20}\textbf{0.139} & 0.238 & \cellcolor{green!20}0.201 & 0.531 & \cellcolor{green!20}0.175 \\
\midrule
\multirow[c]{5}{*}{$\delta_1\uparrow$} 
 & Hamlyn & \textbf{0.351} & 0.256 & 0.264 & 0.247 & 0.229 & \cellcolor{green!20}0.248 & 0.259 & \cellcolor{green!20}\underline{0.336} \\
 & EndoNerf & 0.431 & \cellcolor{green!20}\textbf{0.794} & 0.722 & \cellcolor{green!20}0.737 & 0.720 & \cellcolor{green!20}0.732 & 0.131 & \cellcolor{green!20}\underline{0.791} \\
 & SCARED & 0.469 & \cellcolor{green!20}0.473 & 0.465 & \cellcolor{green!20}\underline{0.478} & \textbf{0.481} & 0.460 & 0.240 & \cellcolor{green!20}0.454 \\
 & StereoMIS & 0.532 & \cellcolor{green!20}\textbf{0.619} & 0.584 & 0.582 & \underline{0.616} & 0.567 & 0.183 & \cellcolor{green!20}0.483 \\
 & StereoKP & 0.508 & \cellcolor{green!20}\underline{0.746} & 0.513 & \cellcolor{green!20}\textbf{0.787} & 0.534 & \cellcolor{green!20}0.620 & 0.227 & \cellcolor{green!20}0.691 \\
\bottomrule

\end{tabular}
\end{table}

\subsection{Experiments}
\label{sec:experiments}

All datasets are preprocessed to ensure metric depth annotations in meters. For datasets lacking direct metric depth (StereoMIS and Hamlyn), disparity maps from rectified stereo pairs are converted to metric depth using a pretrained Unimatch~\cite{xu2023unifying} and camera intrinsics parameters.
All models are fine-tuned exclusively on our synthetic dataset, EndoSynth, for metric depth estimation. Simlarly to prior work~\cite{birkl2023midas,yang2024depth}, we use a scale invariant logarithmic loss~\cite{eigen2014depth} with $\lambda=0.5$ for supervision on metric depth maps. We use a scale invariant logarithmic gradient loss as in~\cite{ranftl2020towards} to improve sharpness. We additionally use a multi-scale edge-aware log smoothness term adapted from optical flow community~\cite{tomasi1998bilateral}. The initial weights for all models were sourced from author-provided checkpoints. All experiments use an Adam optimizer~\cite{kingma2014adam} with $\beta_1=0.9$, $\beta_2=0.99$, and a fixed learning rate of $lr=5 \cdot 10^{-6}$ on all weights of the entire model. We use colour jitter, temperature scaling, motion blur and random smoke from Albumentations~\cite{buslaev2020albumentations} for augmentation during training. The training is stopped after 20 epochs, where an epoch comprises 500 random samples from EndoSynth.

\begin{table}[b]
\tablefontsize

\centering
\caption{Zero-shot \emph{metric} depth performances on each test dataset. Models fine-tuned using EndoSynth only. Mean average Error (MAE) and accuracy at 2cm (Acc@2cm) (\textbf{best}, \underline{second best}).}
\label{tab:eval_metricdepth}
\begin{tabular}{|c|l|c|c|c|c|}
\toprule
 & model & DAv1-F & DAv2-F & EndoDAC-F & Midas-F \\
metric & dataset &  &  &  &  \\
\midrule
\multirow[c]{3}{*}{MAE $\downarrow$} 
 & Hamlyn & 2.02 & \underline{1.86} & 2.70 & \textbf{1.74} \\
 & EndoNerf & 2.68 & \textbf{1.72} & 3.74 & \underline{2.06} \\
 & SCARED & \underline{3.55} & 5.21 & 4.90 & \textbf{2.67} \\
\multirow[c]{2}{*}{(cm)} & StereoMIS & 3.29 & \textbf{2.86} & \underline{3.06} & 3.73 \\
 & StereoKP & \textbf{1.01} & \underline{1.33} & 1.56 & 1.39 \\
\midrule
\multirow[c]{5}{*}{Acc@2cm $\uparrow$} 
 & Hamlyn & 0.57 & \underline{0.60} & 0.42 & \textbf{0.65} \\
 & EndoNerf & 0.22 & \textbf{0.63} & 0.09 & \underline{0.47} \\
 & SCARED & \underline{0.22} & 0.03 & 0.09 & \textbf{0.50} \\
 & StereoMIS & 0.42 & \textbf{0.44} & \underline{0.43} & 0.38 \\
 & StereoKP & \textbf{0.87} & \underline{0.81} & 0.72 & 0.77 \\
\bottomrule
\end{tabular}
\end{table}

\noindent \textbf{Instrument semantic segmentation:} Additionally to depth estimation, we explore the usability of EndoSynth for multi-task learning. We augment the best performing model (DAv2) with an auxiliary decoding branch for surgical instrument semantic segmentation. To achieve this, we modify the last layer of DTPHead~\cite{yang2024depthv2} to output two dense map channels instead of one. The weights are shared until the last 2 convolutional layers. We use binary cross entropy loss supervised on the surgical instrument mask provided in EndoSynth. We train our network end-to-end for both tasks simultaneously. 

\subsection{Evaluation}
\label{sec:evaluation}

All evaluations are performed in a zero-shot manner. We compare models taken off-the-shelf against models after being fine-tuned as described in~\ref{sec:experiments}. Depth estimation evaluation is performed on all 5 unseen real datasets.

\noindent \textbf{Relative depth:} Following most depth model benchmarks~\cite{birkl2023midas,yang2024depth}, we evaluate performance of the models using the absolute relative error $\text{AbsRel} = | d - d^* | d$ and $\delta_1$ (percentage of $\text{max}(d / d^*, d^* / d)$. These metrics are scale-invariant. We rescale the predicted depth map using a median scaling method as in~\cite{cui2024endodac,shao2022self}.

\noindent \textbf{Metric depth:} We evaluate metric performance using the mean absolute error (MAE) in centimetres and an accuracy metric: the percentage of samples with error $\leq$ 2cm (Acc@2cm).
For StereoKP dataset, we take advantage of the fact that the instrument keypoints are annotated,  providing a more accurate and reliable source of ground truth depth information. We therefore evaluate metrics on keypoints depth for this dataset.

\begin{figure}[t]
\includegraphics[width=.99\textwidth]{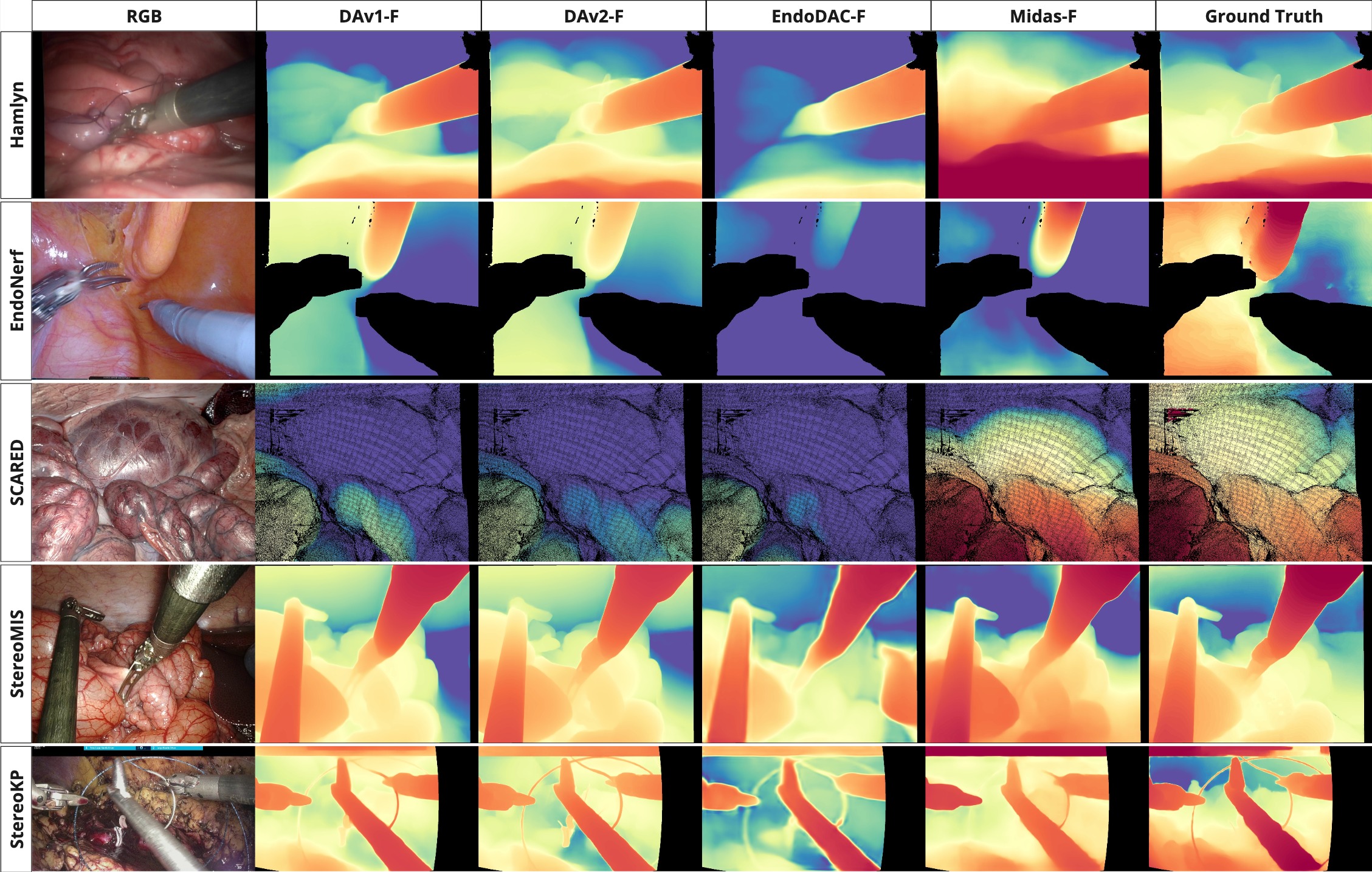}
\caption{Zero-shot monocular \emph{metric} depth. Left-most: RGB, right-most: ground truth metric depth, middle columns: outputs from models fine-tuned exclusively using EndoSynth dataset. \underline{The colour-scale is common to all columns}, corresponding to the ground truth depth range.}
\label{fig:output_samples}
\end{figure}

\section{Results}

\textbf{Zero-shot relative depth estimation:} Table~\ref{tab:eval_reldepth} compares the performance of off-the-shelf models against those fine-tuned on EndoSynth for zero-shot relative depth estimation. Fine-tuning on EndoSynth yields consistent improvements across DAv1, DAv2, and Midas architectures, with notable gains in AbsRel and $\delta_1$ metrics. These results confirm the dataset’s capacity to enhance generalization to unseen surgical environments. We note that models underperform on the SCARED dataset, likely due to its divergent depth ranges compared to EndoSynth and other datasets (Fig.~\ref{fig:endosynth_samples}, right).

\noindent \textbf{Zero-Shot Metric Depth Estimation:} Table~\ref{tab:eval_metricdepth} quantifies metric depth prediction accuracy, reporting mean absolute errors (MAE) below 3cm on four of five evaluated datasets. The DAv2-F variant achieves the strongest overall performance, with over 80\% of StereoKP samples exhibiting errors under 2cm (Acc@2cm metric).

\begin{figure}[t]
    \includegraphics[width=.99\textwidth]{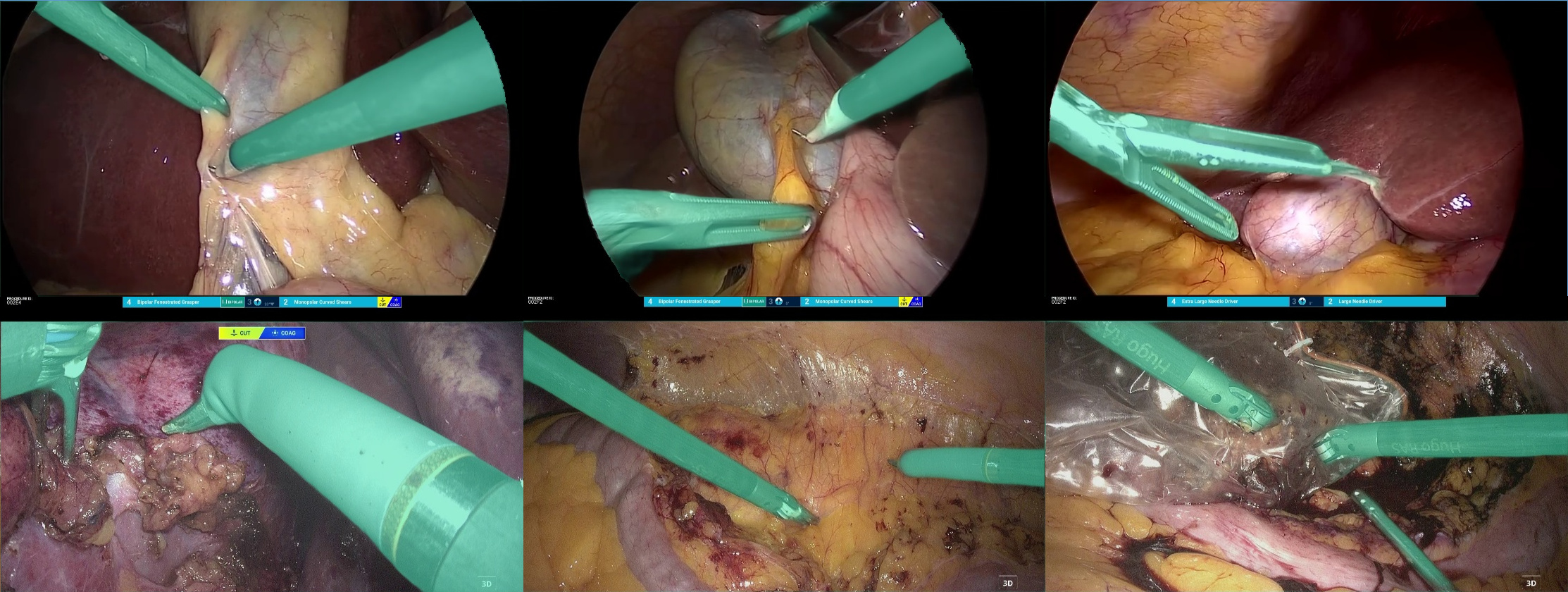}
\caption{Multi-task learning: samples of zero-shot semantic segmentation on CholecSeg8k~\cite{hong2020cholecseg8k} (top) and StereoKP (bottom) using DAv2-F augmented with segmentation head.}
\label{fig:seg}
\end{figure}

\noindent  \textbf{Qualitative results:} Results of \emph{metric} depth maps from all trained models are presented in Fig.~\ref{fig:output_samples}. They demonstrate robust generalisation across most datasets, particularly for DAv2-F and Midas. We note however the under-performance on SCARED, again due to the shift in depth range compared to other datasets. There is also a noticeable loss of granularity for background anatomy, due to EndoSynth’s constrained anatomical diversity. This observation emphasizes the need for expanded synthetic background variability, including broader camera depth ranges and procedural conditions (e.g., tissue deformation, fluid dynamics). DAv2 shows the most detailed outputs for StereoMIS, EndoNerf and StereoKP datasets.

\noindent \textbf{Instrument semantic segmentation:} We evaluate zero-shot semantic segmentation performance on the CholecSeg8k~\cite{hong2020cholecseg8k} and StereoKP datasets. As illustrated in Fig.~\ref{fig:seg}, qualitative results reveal precise instrument delineation. Quantitative analysis yields mean Intersection over Union (mIoU) scores of $0.63$ on CholecSeg8k and $0.83$ on StereoKP. Notably, these results are achieved without any real-world training data—models are fine-tuned exclusively on our synthetic EndoSynth dataset. 

Collectively, these findings validate EndoSynth as a powerful training resource, robust generalization to unseen real surgical environments for the task of depth estimation. Additionally show promising initial results on multi-task learning with segmentation as an auxiliary task. The results also highlight key areas for future refinement in synthetic data generation pipelines.

\section{Conclusion}

This work addresses the critical challenge of \emph{metric} depth estimation in minimally invasive surgical imaging through two primary contributions. First, we introduce EndoSynth, a novel synthetic dataset with high-fidelity endoscopic scenes with Hugo\textsuperscript{TM} RAS instruments positioned in an anatomical abdominal cavity. Each synthetic frame is paired with pixel-wise ground truth metric depth and instrument segmentation masks, providing precise supervision for training depth estimation models. Second, we benchmark state-of-the-art architectures fine-tuned on EndoSynth, demonstrating robust generalization to unseen real endoscopic datasets for the task of metric depth estimation. Our experiments validate the dataset’s utility in bridging the simulation-to-reality gap. We also demonstrate that multi-task learning with a semantic segmentation as secondary task shows impressive performance on unseen real datasets. We release EndoSynth as an open resource, offering the research community a standardized framework for developing and validating surgical depth estimation algorithms under controlled yet clinically relevant conditions.
While EndoSynth advances synthetic data generation for surgical contexts, we acknowledge limitations in anatomical background diversity, which may reduce granularity in predicted depth maps. Future improvements could use compositional augmentation strategies, leveraging the provided segmentation masks to integrate diverse synthetic or real backgrounds (e.g., textures, lighting variations, simulated fluids, or smoke).

By providing a benchmark for surgical depth estimation and a scalable synthetic training platform, this work is a step towards accurate metric depth prediction in robot-assisted interventions. We anticipate that EndoSynth will accelerate progress in surgical vision systems, ultimately improving precision and safety in minimally invasive procedures.
\bibliographystyle{splncs04}
\bibliography{2025_miccai_toussaint_mde}

\end{document}